\documentclass[USenglish,oneside,twocolumn]{article}

\usepackage[utf8]{inputenc}%
\usepackage[big]{dgruyter_NEW}
 
\DOI{foobar}

\cclogo{\includegraphics{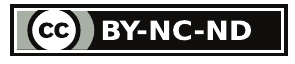}}

\RequirePackage{fix-cm}
\usepackage{natbib}
\usepackage{dblfloatfix}

\makeatletter %
\renewcommand\huge{\@setfontsize\huge{19.9}{26}}
\makeatother

\makeatletter
\def\blfootnote{\xdef\@thefnmark{}\@footnotetext}
\makeatother

\usepackage[utf8]{inputenc} %
\usepackage[T1]{fontenc}    %
\usepackage{hyperref}       %
\usepackage{url}            %
\usepackage{booktabs}       %
\usepackage{amsfonts}       %
\usepackage{nicefrac}       %
\usepackage{microtype}      %
\usepackage{wrapfig}
\usepackage{adjustbox}

\usepackage{microtype}
\usepackage{subfig}

\usepackage{etoolbox}
\makeatletter
\patchcmd\@combinedblfloats{\box\@outputbox}{\unvbox\@outputbox}{}
\makeatother

\usepackage[algoruled,vlined,nofillcomment]{algorithm2e}

\usepackage{hyperref}
\usepackage{amsmath}

\usepackage{amsthm}
\usepackage[textsize=footnotesize,color=green!40]{todonotes}
\usepackage{amssymb}
\usepackage{cleveref}
\usepackage{etoolbox} %
\usepackage{floatrow}
\newfloatcommand{capbtabbox}{table}[][\FBwidth]

\usepackage[font=small]{caption}

\Crefname{equation}{Eq.}{Eqs.}%
\Crefname{figure}{Fig.}{Figs.}%
\Crefname{section}{Sec.}{Secs.}%
\Crefname{algorithm}{Alg.}{Algs.}%
\Crefname{theorem}{Thm.}{Thms.}%
\Crefname{appendix}{Appx.}{Appxs.}

\newcommand{\R}{\mathbb{R}}

\renewcommand{\Pr}{\mathbb{P}}

\newcommand{\clip}{\mathsf{clip}}
\newcommand{\cN}{\mathcal{N}}
\newcommand{\cA}{\mathcal{A}}
\newcommand{\cP}{\mathcal{P}}
\newcommand{\Z}{\mathcal{Z}}
\newcommand{\X}{\mathcal{X}}
\newcommand{\Y}{\mathcal{Y}}
\newcommand{\W}{\mathcal{W}}
\newcommand{\V}{\mathcal{V}}
\newcommand{\privor}{\mathsf{P}}
\newcommand{\utilor}{\mathsf{U}}
\newcommand{\vol}{\mathsf{HV}}
\newcommand{\pareto}{\mathcal{PF}}
\newcommand{\Lap}{\mathsf{Lap}}
\newcommand{\norm}[1]{\| #1 \|}

\newcommand{\Dcal}{\mathcal{D}}
\newcommand{\Ical}{\mathcal{I}}

\newcommand{\antiideal}{v^\dagger}

\newtheorem{theorem}{Theorem}

\newtheorem{definition}[theorem]{Definition}
\theoremstyle{definition}
\newtheorem{remark}{Remark}

\usepackage{acronym}

\makeatletter
\newif\ifAC@uppercase@first%
\def\Aclp#1{\AC@uppercase@firsttrue\aclp{#1}\AC@uppercase@firstfalse}%
\def\AC@aclp#1{%
  \ifcsname fn@#1@PL\endcsname%
    \ifAC@uppercase@first%
      \expandafter\expandafter\expandafter\MakeUppercase\csname fn@#1@PL\endcsname%
    \else%
      \csname fn@#1@PL\endcsname%
    \fi%
  \else%
    \AC@acl{#1}s%
  \fi%
}%
\def\Acp#1{\AC@uppercase@firsttrue\acp{#1}\AC@uppercase@firstfalse}%
\def\AC@acp#1{%
  \ifcsname fn@#1@PL\endcsname%
    \ifAC@uppercase@first%
      \expandafter\expandafter\expandafter\MakeUppercase\csname fn@#1@PL\endcsname%
    \else%
      \csname fn@#1@PL\endcsname%
    \fi%
  \else%
    \AC@ac{#1}s%
  \fi%
}%
\def\Acfp#1{\AC@uppercase@firsttrue\acfp{#1}\AC@uppercase@firstfalse}%
\def\AC@acfp#1{%
  \ifcsname fn@#1@PL\endcsname%
    \ifAC@uppercase@first%
      \expandafter\expandafter\expandafter\MakeUppercase\csname fn@#1@PL\endcsname%
    \else%
      \csname fn@#1@PL\endcsname%
    \fi%
  \else%
    \AC@acf{#1}s%
  \fi%
}%
\def\Acsp#1{\AC@uppercase@firsttrue\acsp{#1}\AC@uppercase@firstfalse}%
\def\AC@acsp#1{%
  \ifcsname fn@#1@PL\endcsname%
    \ifAC@uppercase@first%
      \expandafter\expandafter\expandafter\MakeUppercase\csname fn@#1@PL\endcsname%
    \else%
      \csname fn@#1@PL\endcsname%
    \fi%
  \else%
    \AC@acs{#1}s%
  \fi%
}%
\edef\AC@uppercase@write{\string\ifAC@uppercase@first\string\expandafter\string\MakeUppercase\string\fi\space}%
\def\AC@acrodef#1[#2]#3{%
  \@bsphack%
  \protected@write\@auxout{}{%
    \string\newacro{#1}[#2]{\AC@uppercase@write #3}%
  }\@esphack%
}%
\def\Acl#1{\AC@uppercase@firsttrue\acl{#1}\AC@uppercase@firstfalse}
\def\Acf#1{\AC@uppercase@firsttrue\acf{#1}\AC@uppercase@firstfalse}
\def\Ac#1{\AC@uppercase@firsttrue\ac{#1}\AC@uppercase@firstfalse}
\def\Acs#1{\AC@uppercase@firsttrue\acs{#1}\AC@uppercase@firstfalse}
\robustify\Aclp
\robustify\Acfp
\robustify\Acp
\robustify\Acsp
\robustify\Acl
\robustify\Acf
\robustify\Ac
\robustify\Acs
\makeatother

\acrodef{BO}{Bayesian optimization}
\acrodef{HPO}{hyperparameter optimization}
\acrodef{DP}{differential privacy}
\acrodef{SGD}{stochastic gradient descent}
\acrodef{GP}{Gaussian process}
\acrodef{LogReg}{logistic regression}
\acrodef{SVM}{support vector machine}
\acrodef{MLP}{multilayer perceptron}
\acrodef{PoI}{\emph{probability of improvement}}
\acrodef{HVPoI}{\emph{hypervolume-based} \ac{PoI}}

\newcommand{\gp}{\ac{GP}}
\newcommand{\DPareto}{\textsc{DPareto}}

\newcommand{\MNIST}{{\scshape mnist}}
\newcommand{\Adult}{{\scshape adult}}

\usepackage{siunitx}
\begin{document}

  \author[1]{Brendan Avent}

  \author[2]{Javier Gonz\'{a}lez}

  \author[3]{Tom Diethe}

  \author[4]{Andrei Paleyes}

  \author[5]{Borja Balle}

  \affil[1]{University of Southern California$^\dagger$, E-mail: bavent@usc.edu}

  \affil[2]{Now at Microsoft Research$^\dagger$, E-mail: Gonzalez.Javier@microsoft.com}

  \affil[3]{Amazon Research Cambridge, E-mail: tdiethe@amazon.com}

  \affil[4]{Now at University of Cambridge$^\dagger$, E-mail: ap2169@cam.ac.uk}

  \affil[5]{Now at DeepMind$^\dagger$, E-mail: borja.balle@gmail.com}

  \title{\huge Automatic Discovery of Privacy--Utility Pareto~Fronts}

  \runningtitle{Automatic Discovery of Privacy--Utility Pareto Fronts}

\begin{abstract}
{Differential privacy is a mathematical framework for privacy-preserving data analysis.
Changing the hyperparameters of a differentially private algorithm allows one to trade off privacy and utility in a principled way.
Quantifying this trade-off in advance is essential to decision-makers tasked with deciding how much privacy can be provided in a particular application while maintaining acceptable utility.
Analytical utility guarantees offer a rigorous tool to reason about this trade-off, but are generally only available for relatively simple problems.
For more complex tasks, such as training neural networks under differential privacy, the utility achieved by a given algorithm can only be measured empirically.
This paper presents a Bayesian optimization methodology for efficiently characterizing the privacy--utility trade-off of any differentially private algorithm using only empirical measurements of its utility.
The versatility of our method is illustrated on a number of machine learning tasks involving multiple models, optimizers, and datasets.}
\end{abstract}


  \journalname{Proceedings on Privacy Enhancing Technologies}
\DOI{Editor to enter DOI}
  \startpage{1}

\maketitle

\section{Introduction}\label{sec:intro}
\blfootnote{This paper appears in the 2020 Proceedings on Privacy Enhancing Technologies.}
\blfootnote{$^\dagger$ Work done while at Amazon Research Cambridge.}
\Ac{DP} \cite{dwork2006calibrating} is the de-facto standard for privacy-preserving data analysis, including the training of machine learning models using sensitive data.
The strength of \ac{DP} comes from its use of randomness to hide the contribution of any individual's data from an adversary with access to \emph{arbitrary side knowledge}.
The price of \ac{DP} is a loss in utility caused by the need to inject
noise into computations. 
Quantifying the trade-off between privacy and utility is a central topic in the literature on differential privacy.
Formal analysis of such trade-offs lead to algorithms achieving a pre-specified level privacy with minimal utility reduction, or, conversely, an a priori acceptable level of utility with maximal privacy.
Since the choice of privacy level is generally regarded as a policy decision \cite{1021596}, %
this quantification is essential to decision-makers tasked with balancing utility and privacy in real-world deployments \cite{abowd2018economic}.

However, analytical analyses of the privacy--utility trade-off are only available for relatively simple problems amenable to mathematical treatment, and cannot be conducted for most problems of practical interest.
Further, differentially private algorithms have more hyperparameters than their non-private counterparts, most of which affect both privacy and utility.
In practice, tuning these hyperparameters to achieve an optimal privacy--utility trade-off can be an arduous task, especially when the utility guarantees are loose or unavailable.
In this paper we develop a Bayesian optimization approach for \emph{empirically} characterizing any differentially private algorithm's privacy--utility trade-off via principled, computationally efficient hyperparameter tuning.

A canonical application of our methods is differentially private deep learning.
Differentially private stochastic optimization has been employed to train feed-forward \cite{abadi2016deep}, convolutional \cite{DBLP:journals/corr/abs-1802-08232}, and recurrent \cite{brendan2018learning} neural networks, showing that reasonable accuracies can be achieved when selecting hyperparameters carefully.
These works rely on a differentially private \emph{gradient perturbation} technique, which clips and adds noise to gradient computations, while keeping track of the privacy loss incurred.
However, these results do not provide actionable information regarding the privacy--utility trade-off of the proposed models.
For example, private stochastic optimization methods can obtain the same level of privacy in different ways (e.g.\ by increasing the noise variance and reducing the clipping norm, or \emph{vice-versa}), and it is not generally clear what combinations of these changes yield the best possible utility for a fixed privacy level.
Furthermore, increasing the number of hyperparameters makes exhaustive hyperparameter optimization prohibitively expensive.

The goal of this paper is to provide a computationally efficient methodology to this problem by using Bayesian optimization to non-privately estimate the privacy--utility \emph{Pareto front} of a given differentially private algorithm.
The Pareto fronts obtained by our method can be used to select hyperparameter settings of the optimal operating points of any differentially private technique, enabling decision-makers to take informed actions when balancing the privacy--utility trade-off of an algorithm before deployment.
This is in line with the approach taken by the U.S.\ Census Bureau to calibrate the level of \ac{DP} that will be used when releasing the results of the upcoming 2020 census \cite{garfinkel2018issues,abowd2018economic,fall-csac}.

Our contributions are: (1) Characterizing the privacy--utility trade-off of a hyperparameterized algorithm as the problem of learning a Pareto front on the privacy vs.\ utility plane (\Cref{sec:setup}).
(2) Designing \DPareto{}, an algorithm that leverages multi-objective Bayesian optimization techniques for learning the privacy--utility Pareto front of any differentially private algorithm (\Cref{sec:dpareto}).
(3) Instantiating and experimentally evaluating our framework for the case of differentially private stochastic optimization on a variety of learning tasks involving multiple models, optimizers, and datasets (\Cref{sec:experiments}).
Finally, important and challenging extensions to this work are proposed (\Cref{sec:extensions}) and closely-related work is reviewed (\Cref{sec:related}).

\vspace*{-0ex}
\section{The Privacy--Utility Pareto Front}\label{sec:setup}

This section provides an abstract formulation of the problem we want to address.
We start by introducing some basic notation and recalling the definition of \acl{DP}.
We then formalize the task of quantifying the privacy--utility trade-off using the notion of a Pareto front.
Finally, we conclude by illustrating these concepts using classic examples from both machine learning as well as \acl{DP}.

\subsection{General Setup}
Let $A : \Z^n \to \W$ be a \emph{randomized algorithm} that takes as input a tuple containing $n$ records from $\Z$ and outputs a value in some set $\W$. %
Differential privacy formalizes the idea that $A$ preserves the privacy of its inputs when the output distribution is stable under changes in one input.

\begin{definition}[\cite{dwork2006calibrating,DBLP:conf/icalp/Dwork06}]
Given $\varepsilon \geq 0$ and $\delta \in [0,1]$, we say algorithm $A$ is
$(\varepsilon,\delta)$-DP if for any pair of inputs $z, z'$ differing in a single coordinate we have\footnote{Smaller values of $\varepsilon$ and $\delta$ yield more private algorithms.}
\begin{align*}
\sup_{E \subseteq \W} \left( \Pr[A(z) \in E] - e^{\varepsilon} \Pr[A(z') \in E] \right) \leq \delta \enspace.
\end{align*}
\end{definition}

To analyze the trade-off between utility and privacy for a given problem, we consider a \emph{parametrized} family of algorithms $\cA = \{ A_\lambda: \Z^n \to \W\}$. %
Here, $\lambda \in \Lambda$ indexes the possible choices of hyperparameters, so $\cA$ can be interpreted as the set of all possible algorithm configurations for solving a given task.
For example, in the context of a machine learning application, the family $\cA$ consists of a set of learning algorithms which take as input a training dataset $z = (z_1, \ldots, z_n)$ containing $n$ example-label pairs $z_i = (x_i, y_i) \in \Z = \X \times \Y$ and produce as output the parameters $w \in \W \subseteq \R^d$ of a predictive model.
It is clear that in this context different choices for the hyperparameters might yield different utilities.
We further assume each configuration $A_\lambda$ of the algorithm satisfies \ac{DP} with potentially distinct privacy parameters.

To capture the privacy--utility trade-off across $\cA$ we introduce two oracles to model the effect of hyperparameter changes on the privacy and utility of $A_\lambda$.
A \emph{privacy oracle} is a function $\privor_{\delta} : \Lambda \to [0,+\infty]$ that given a choice of hyperparameters $\lambda$ returns a value $\varepsilon = \privor_{\delta}(\lambda)$ such that $A_\lambda$ satisfies $(\varepsilon, \delta)$-\ac{DP} for a given $\delta$.
An instance-specific \emph{utility oracle} is a function $\utilor_{z} : \Lambda \to [0,1]$ that given a choice of hyperparameters $\lambda$ returns some measure of the utility\footnote{Due to the broad applicability of \ac{DP}, concrete utility measures are generally defined on a per-problem basis. Here we use the conventions that $\utilor_{z}$ is bounded and that larger utility is better.} %
of the output distribution of $A_{\lambda}(z)$.
These oracles allow us to condense everything about our problem in the tuple $(\Lambda, \privor_{\delta}, \utilor_{z})$.
Given these three objects, our goal is to find hyperparameter settings for $A_\lambda$ that simultaneously achieve maximal privacy and utility on a given input $z$. Next we will formalize this goal using the concept of Pareto front, but we first provide remarks about the definition of our oracles.

\begin{remark}[Privacy Oracle]
Parametrizing the privacy oracle $\privor_{\delta}$ in terms of a fixed $\delta$ stems from the convention that $\varepsilon$ is considered the most important privacy parameter\footnote{
This choice comes without loss of generality since there is a connection between the two parameters guaranteeing the existence of a valid $\varepsilon$ for any valid $\delta$ \cite{DBLP:conf/nips/BalleBG18}.}, whereas $\delta$ is chosen to be a negligibly small value ($\delta \ll 1/n$).
This choice is also aligned with recent uses of \ac{DP} in machine learning where the privacy analysis is conducted under the framework of R{\'e}nyi \ac{DP} \cite{mironov2017renyi} and the reported privacy is obtained a posteriori by converting the guarantees to standard $(\varepsilon,\delta)$-DP for some fixed $\delta$ \cite{abadi2016deep,geumlek2017renyi,brendan2018learning,feldman2018privacy,wang-balle-kasiviswanathan18}.
In particular, in our experiments with gradient perturbation for stochastic optimization methods (\Cref{sec:experiments}), we implement the privacy oracle using the moments accountant technique proposed by Abadi et al.~\cite{abadi2016deep} coupled with the tight bounds provided by Wang et al.~\cite{wang-balle-kasiviswanathan18} for R{\'e}nyi \ac{DP} amplification by subsampling without replacement.
More generally, privacy oracles can take the form of analytic formulas or numerical optimized calculations, but future advances in empirical or black-box evaluation of \ac{DP} guarantees could also play the role of privacy oracles.
\end{remark}

\begin{remark}[Utility Oracle]
Parametrizing the utility oracle $\utilor_{z}$ by a fixed input is a choice justified by the type of applications we tackle in our experiments (cf.\ \Cref{sec:experiments}). Other applications might require variations which our framework can easily accommodate by extending the definition of the utility oracle.
We also stress that since the algorithms in $\cA$ are randomized, the utility $\utilor_z(\lambda)$ is a property of the output \emph{distribution} of $A_\lambda(z)$. This means that in practice we might have to implement the oracle approximately, e.g.\ through sampling.
In particular, in our experiments we use a test set to measure the utility of a hyperparameter setting by running $A_{\lambda}(z)$ a fixed number of times $R$ to obtain model parameters $w_1, \ldots, w_R$, and then let $\utilor_z(\lambda)$ be the average accuracy of the models on the test set.
\end{remark}

The Pareto front of a collection of points $\Gamma \subset \R^p$ contains all the points in $\Gamma$ where none of the coordinates can be decreased further without increasing some of the other coordinates (while remaining inside $\Gamma$).

\begin{definition}[Pareto Front]
Let $\Gamma \subset \R^p$ and $u, v \in \Gamma$. We say that $u$ \emph{dominates} $v$ if $u_i \leq v_i$ for all $i \in [p]$, and we write $u \preceq v$. The \emph{Pareto front} of $\Gamma$ is the set of all non-dominated points $\pareto(\Gamma) = \{ u \in \Gamma \;|\; v \not\preceq u,\; \forall\, v \in \Gamma \setminus \{u\} \}$.
\end{definition}

According to this definition, given a privacy--utility trade-off problem of the form $(\Lambda, \privor_{\delta}, \utilor_{z})$, we are interested in finding the Pareto front $\pareto(\Gamma)$ of the $2$-dimensional set\footnote{The use of $1-\utilor_{z}(\lambda)$ for the utility coordinate is for notational consistency, since we use the convention that the points in the Pareto front are those that minimize each individual dimension.} $\Gamma = \{ (\privor_{\delta}(\lambda), 1-\utilor_{z}(\lambda)) \; | \; \lambda \in \Lambda \}$.
If able to fully characterize this Pareto front, a decision-maker looking to deploy \ac{DP} would have all the necessary information to make an informed decision about how to trade-off privacy and utility in their application.

\vspace*{-2ex}
\paragraph*{Threat Model Discussion}
In the idealized setting presented above, the desired output is the Pareto front $\pareto(\Gamma)$ which depends on $z$ through the utility oracle; this is also the case for the Bayesian optimization algorithm for approximating the Pareto front presented in \Cref{sec:dpareto}.
This warrants a discussion about what threat model is appropriate here.

\Ac{DP} guarantees that an adversary observing the output $w = A_{\lambda}(z)$ will not be able to infer too much about any individual record in $z$.
The (central) threat model for \ac{DP} assumes that $z$ is owned by a trusted curator that is responsible for running the algorithm and releasing its output to the world.
However, the framework described above does not attempt to prevent information about $z$ from being exposed by the Pareto front.
This is because our methodology is only meant to provide a substitute for using closed-form utility guarantees when selecting hyperparameters for a given \ac{DP} algorithm \emph{before its deployment}.
Accordingly, throughout this work we assume the Pareto fronts obtained with our method are only revealed to a small set of trusted individuals, which is the usual scenario in an industrial context.
Privatization of the estimated Pareto front would remove the need for this assumption, and is discussed in Sec.~\ref{sec:extensions} as a useful extension of this work.

An alternative approach is to assume the existence of a \emph{public} dataset $z_0$ following a similar distribution to the private dataset $z$ on which we would like to run the algorithm. Then we can use $z_0$ to compute the Pareto front of the algorithm, select hyperparameters $\lambda^*$ achieving a desired privacy--utility trade-off, and release the output of $A_{\lambda^*}(z)$.
In particular, this is the threat model used by the U.S.\ Census Bureau to tune the parameters for their use of \ac{DP} in the context of the 2020 census (see \Cref{sec:related} for more details).

\subsection{Two Illustrative Examples} \label{sec:illustrations}
To concretely illustrate the oracles and Pareto front concept, we consider two distinct examples: private logistic regression and the sparse vector technique.
Both examples are computationally light, and thus admit computation of near-exact Pareto fronts via a fine-grained grid-search on a low-dimensional hyperparameter space; for brevity, we subsequently refer to these as the ``exact'' or ``true'' Pareto fronts.

\begin{figure*}[h]
\begin{center}
\includegraphics[clip,width=.75\textwidth]{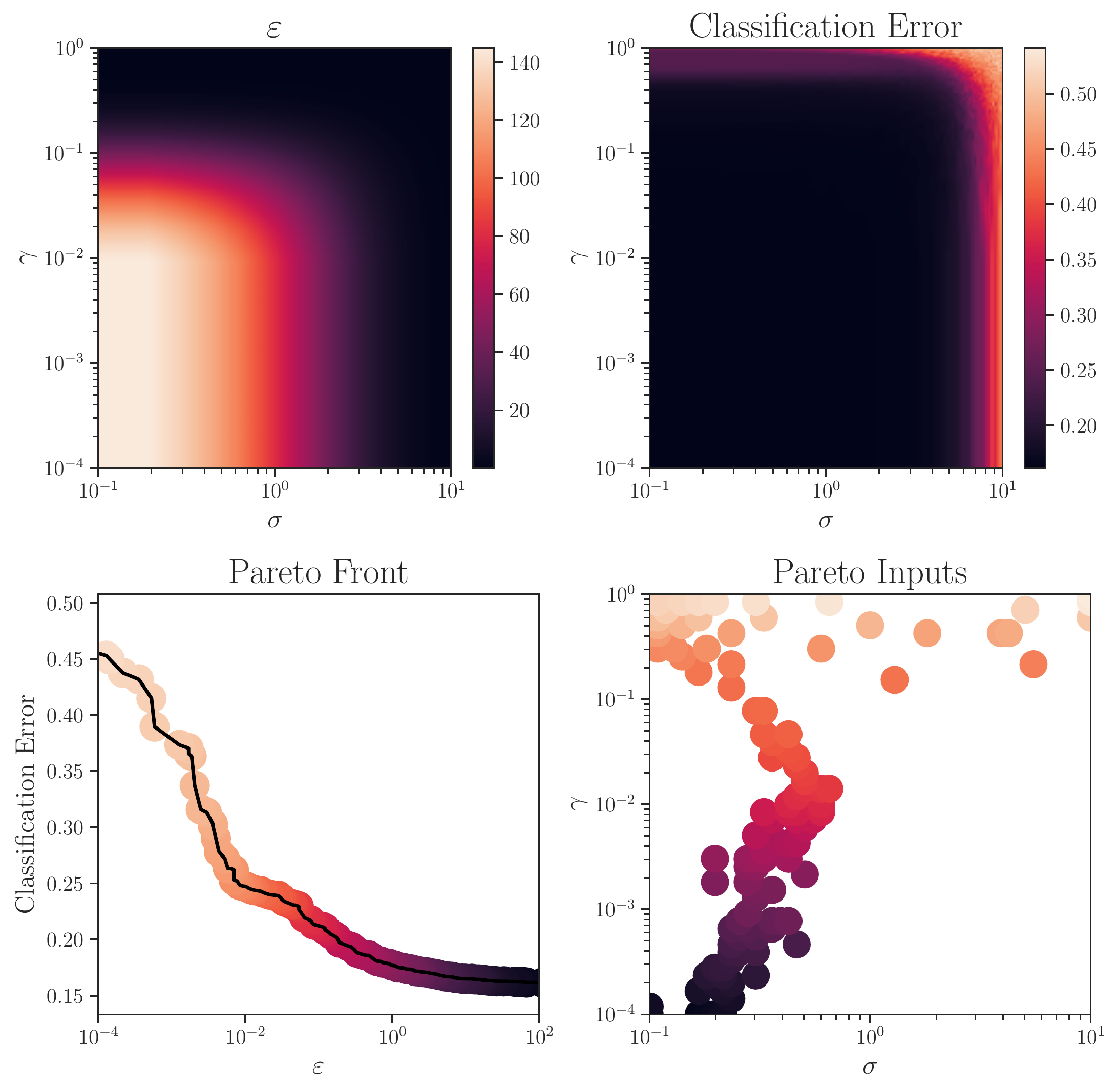}
\end{center}
\vspace*{-3ex}
\caption{\emph{Top:} Values returned by the privacy and utility oracles across a range of hyperparameters in the private logistic regression example. \emph{Bottom:} The Pareto front and its corresponding set of input points.}\label{fig:psgd_exact}
\end{figure*}

\vspace*{-2ex}
\paragraph*{Private Logistic Regression}
Here, we consider a simple private logistic regression model with $\ell_2$ regularization trained on the Adult dataset \cite{kohavi1996scaling}.
The model is privatized by training with mini-batched projected SGD, then applying a Gaussian perturbation at the output using the method from \cite[Algorithm 2]{wu2017bolt} with default parameters\footnote{These are the smoothness, Lipschitz and strong convexity parameters of the loss, and the learning rate.}.
The only hyperparameters tuned in this example are the regularization $\gamma$ and the noise standard deviation $\sigma$, while the rest are fixed\footnote{Mini-batch size $m=1$ and number of epochs $T = 10$.}.
Note that both hyperparameters affect privacy and accuracy in this case.
To implement the privacy oracle we compute the global sensitivity according to \cite[Algorithm 2]{wu2017bolt} and find the $\varepsilon$ for a fixed $\delta = 10^{-6}$ using the exact analysis of the Gaussian mechanism provided in \cite{DBLP:conf/icml/BalleW18}.
To implement the utility oracle we evaluate the accuracy of the model on the test set, averaging over 50 runs for each setting of the hyperparameters.
To obtain the exact Pareto front for this problem, we perform a fine grid search over $\gamma \in [10^{-4},10^0]$ and $\sigma \in [10^{-1},10^1]$.
The Pareto front and its corresponding hyperparameter settings are displayed in \Cref{fig:psgd_exact}, along with the values returned by the privacy and utility oracles across the entire range of hyperparameters.

\vspace*{-2ex}
\paragraph*{Sparse Vector Technique}
The \emph{sparse vector technique} (SVT) \cite{dwork2009complexity} is an algorithm to privately run $m$ queries against a fixed sensitive database and release under \ac{DP} the indices of those queries which exceed a certain threshold.
The naming of the algorithm reflects the fact that it is specifically designed to have good accuracy when only a small number of queries are expected to be above the threshold.
The algorithm has found applications in a number of problems, and several variants of it have been proposed~\cite{lyu2017understanding}.

\Cref{alg:SVT} details our construction of a non-interactive version of the algorithm proposed in \cite[Alg. 7]{lyu2017understanding}.
Unlike the usual SVT that is parametrized by the target privacy $\varepsilon$, our construction takes as input a total noise level $b$ and is tailored to answer $m$ binary queries $q_i : \Z^n \to \{0,1\}$ with sensitivity $\Delta = 1$ and fixed threshold $T = 1/2$.
The privacy and utility of the algorithm are controlled by the noise level $b$ and the bound $C$ on the number of answers; increasing $b$ or decreasing $C$ yields a more private but less accurate algorithm.
This noise level is split across two parameters $b_1$ and $b_2$ controlling how much noise is added to the threshold and to the query answers respectively\footnote{The split used by the algorithm is based on the privacy budget allocation suggested in \cite[Section 4.2]{lyu2017understanding}.}.
Privacy analysis of \Cref{alg:SVT} yields the following closed-form privacy oracle for our algorithm: $\privor_0 = (1 + (2C)^{1/3}) (1 + (2C)^{2/3}) b^{-1}$ (refer to \Cref{sec:svt-proof} for proof).

\begin{algorithm}%
\DontPrintSemicolon
\SetKwInput{KwHP}{Hyperparameters}
\caption{Sparse Vector Technique}\label{alg:SVT}
\KwIn{dataset $z$, queries $q_1, \ldots, q_m$}
\KwHP{noise $b$, bound $C$}
$c \leftarrow 0$, $w \leftarrow (0,\ldots,0) \in \{0,1\}^m$\;
$b_1 \leftarrow b / (1 + (2 C)^{1/3})$, $b_2 \leftarrow b - b_1$, $\rho \leftarrow \Lap(b_1)$\;
\For{$i \in [m]$}{
$\nu \leftarrow \Lap(b_2)$\;
\If{$q_i(z) + \nu \geq \frac{1}{2} + \rho$}{
$w_i \leftarrow 1$, $c \leftarrow c + 1$\;
\lIf{$c \geq C$}{\KwRet{$w$}	}
}
}
\KwRet{$w$}
\end{algorithm}

As a utility oracle, we use the \emph{$F_1$-score} between the vector of true answers $(q_1(z), \ldots, q_m(z))$ and the vector $w$ returned by the algorithm.
This measures how well the algorithm identifies the support of the queries that return $1$, while penalizing both for false positives and false negatives.
This is again different from the usual utility analyses of SVT algorithms, which focus on providing an interval around the threshold outside which the output is guaranteed to have no false positives or false negatives \cite{dwork2014algorithmic}.
Our measure of utility is more fine-grained and relevant for practical applications, although to the best of our knowledge no theoretical analysis of the utility of the SVT in terms of $F_1$-score is available in the literature.

\begin{figure*}[h]
\begin{center}
\includegraphics[clip,width=.75\textwidth]{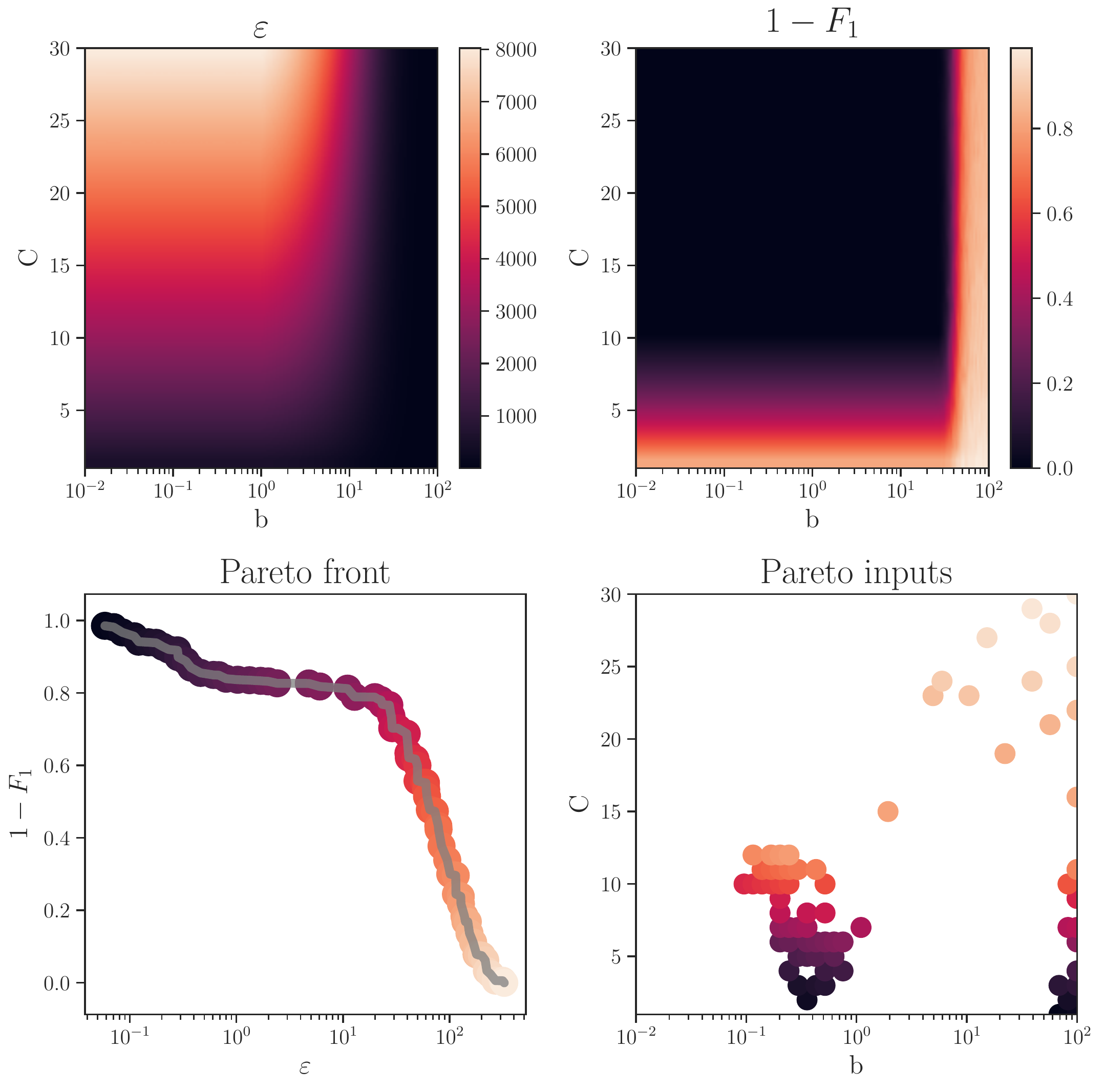}
\end{center}
\vspace*{-3ex}
\caption{\emph{Top:} Values returned by the privacy and utility oracles across a range of hyperparameters in the SVT example. \emph{Bottom:} The Pareto front and its corresponding set of input points.}\label{fig:sparse_vector_exact}
\vspace*{-1ex}
\end{figure*}

In this example, we set $m = 100$ and pick queries at random such that exactly $10$ of them return a $1$.
Since the utility of the algorithm is sensitive to the query order, we evaluate the utility oracle by running the algorithm $50$ times with a random query order and compute the average $F_1$-score.
The Pareto front and its corresponding hyperparameter settings are displayed in \Cref{fig:sparse_vector_exact}, along with the values returned by the privacy and utility oracles across the entire range of hyperparameters.

\section{\DPareto{}: Learning the Pareto Front}\label{sec:dpareto}

This section starts by recalling the basic ideas behind multi-objective \acl{BO}.
We then describe our proposed methodology to efficiently learn the privacy--utility Pareto front.
Finally, we revisit the private logistic regression and SVT examples to illustrate our method.

\vspace*{-2ex}
\subsection{Multi-objective Bayesian Optimization}
\ac{BO} \cite{movckus1975bayesian} is a strategy for sequential decision making useful for optimizing expensive-to-evaluate black-box objective functions. It has become increasingly relevant in machine learning due to its success in the optimization of model hyperparameters \cite{snoek2012practical,jenatton17}. 
In its standard form, \ac{BO} is used to find the minimum of an objective function $f(\lambda)$ on some subset $\Lambda \subseteq \R^p$ of a Euclidean space of moderate dimension. It works by generating a sequence of evaluations of the objective at locations $\lambda_1,\dots,\lambda_k$, which is done by (i) building a surrogate model of the objective function using the current data and (ii) applying a pre-specified criterion to select a new location $\lambda_{k+1}$ based on the model until a budget is exhausted.
In the single-objective case, a common choice is to select the location that, in expectation under the model, gives the best improvement to the current estimate~\cite{movckus1975bayesian}.

In this work, we use BO for learning the privacy--utility Pareto front. When used in multi-objective problems, BO aims to learn a Pareto front with a minimal number of evaluations, which makes it an appealing tool in cases where evaluating the objectives is expensive. Although in this paper we only work with two objective functions, we detail here the general case of minimizing $p$ objectives $f_1,\dots, f_p$ simultaneously.
This generalization could be used, for instance, to introduce the running time of the algorithm as a third objective to be traded off against privacy and utility.

Let $\lambda_1,\dots,\lambda_k$ be a set of locations in $\Lambda$ and denote by $\V =\{v_1,\dots,v_k\}$ the set such that each $v_i \in \mathbb{R}^{p}$ is the vector $(f_1(\lambda_i), \ldots, f_p(\lambda_i))$.
In a nutshell, BO works by iterating over the following: 

\begin{list}{{\arabic{enumi}.}}{\usecounter{enumi}
\setlength{\leftmargin}{19pt}
\setlength{\listparindent}{-1pt}
\setlength{\parsep}{-1pt}}
\item Fit a surrogate model of the objectives $f_1(\lambda) \dots, f_p(\lambda)$ using the available dataset $\Dcal = \{(\lambda_i, v_i) \}_{i=1}^k$. The standard approach is to use a \ac{GP} \cite{Rasmussen:2005}. 
\item For each objective $f_j$ calculate the predictive distribution over $\lambda \in \Lambda$ using the surrogate model. If \acp{GP} are used, the predictive distribution of each output can be fully characterized by their mean $m_j(\lambda)$ and variance $s^2_j(\lambda)$ functions, which can be computed in closed form.  
\item Use the posterior distribution of the surrogate model to form an acquisition function $\alpha(\lambda; \Ical)$, where $\Ical$ represents the %
dataset $\Dcal$ and the \ac{GP} posterior %
conditioned on $\Dcal$.
\item Collect the next evaluation point $\lambda_{k+1}$ at the (numerically estimated) global maximum of $\alpha(\lambda; \Ical)$.
\end{list}
The process is repeated until the budget to collect new locations is over.

There are two key aspects of any BO method: the surrogate model of the objectives and the acquisition function $\alpha(\lambda; \Ical)$.
In this work, we use independent \acp{GP} as the surrogate models for each objective; however, generalizations with multi-output \acp{GP} \cite{Alvarez:2012} are possible.%

\vspace*{-2ex}
\paragraph*{Acquisition with Pareto Front Hypervolume}
Next we define an acquisition criterion $\alpha(\lambda; \Ical)$ useful to collect new points when learning the Pareto front.
Let $\cP = \pareto(\V)$ be the Pareto front computed with the objective evaluations in $\Ical$ and let $\antiideal \in \R^p$ be some chosen ``anti-ideal'' point\footnote{The anti-ideal point must be dominated by all points in $\mathcal{PF}(\V)$. In the private logistic regression example, this could correspond to the worst-case $\epsilon$ and worst-case classification error. See Couckuyt et al.~\cite{Couckuyt:2014} for further details.}.
To measure the relative merit of different Pareto fronts we use the \emph{hypervolume} $\vol_{\antiideal}(\cP)$ of the region dominated by the Pareto front $\cP$ bounded by the anti-ideal point. Mathematically this can be expressed as $\vol_{\antiideal}(\cP) = \mu(\{ v \in \R^p \;|\; v \preceq \antiideal, \; \exists u \in \cP \; u \preceq v \})$,
where $\mu$ denotes the standard Lebesgue measure on $\R^p$.
Henceforth we assume the anti-ideal point is fixed and drop it from our notation.

Larger hypervolume therefore implies that points in the Pareto front are closer to the ideal point $(0,0)$.
Thus, $\vol(\pareto(\V))$ provides a way to measure the quality of the Pareto front obtained from the data in $\V$.  
Furthermore, hypervolume can be used to design acquisition functions for selecting hyperparameters that will improve the Pareto front.
Start by defining the increment in the hypervolume %
given a new point $v \in \R^p$: 
$
\Delta_{\pareto}(v) = \vol(\pareto(\V \cup \{v\})) - \vol(\pareto(\V)). %
$
This quantity is positive only if $v$ lies in the set $\tilde{\Gamma}$ of points non-dominated by $\pareto(\V)$.
Therefore, the \ac{PoI} over the current Pareto front when selecting a new hyperparameter $\lambda$ can be computed using the model trained on $\Ical$ as follows:
\begin{align}\label{eq:probability_improvement}
\mathsf{PoI}(\lambda) &= \Pr[ (f_1(\lambda), \ldots, f_p(\lambda)) \in \tilde{\Gamma} \;|\; \Ical ] \\
&= \int_{v \in \tilde{\Gamma} } \prod_{j=1}^p \phi_j(\lambda; v_j)  d v_j \enspace,
\end{align}
where $\phi_j(\lambda; \cdot)$
is the predictive Gaussian density for $f_j$ with mean $m_j(\lambda)$ and variance $s^2_j(\lambda)$.

The $\mathsf{PoI}(\lambda)$ function (\ref{eq:probability_improvement})
accounts for the probability that a given $\lambda \in \Lambda$ has to improve the Pareto front, and it can be used as a criterion to select new points.
However, in this work, we opt for the \ac{HVPoI} due to its superior computational and practical properties \cite{Couckuyt:2014}.
The \ac{HVPoI} is given by
\begin{equation}\label{eq:hvpoi}
\alpha(\lambda; \Ical) = \Delta_{\mathcal{PF}} (m(\lambda)) \cdot \mathsf{PoI}(\lambda) \enspace,
\end{equation}
where $m(\lambda) = (m_1(\lambda), \dots, m_p(\lambda))$. This acquisition weights the probability of improving the Pareto front with a measure of how much improvement is expected, computed using the GP means of the outputs. The \ac{HVPoI} has been shown to work well in practice and efficient implementations exist \cite{GPflowOpt2017}.

\subsection{The \DPareto~Algorithm}
The main optimization loop of \DPareto{} is shown in \Cref{alg:main}.
It combines the two ingredients sketched so far: \acp{GP} for surrogate modeling of the objective oracles, and \ac{HVPoI} as the acquisition function to select new hyperparameters.
The basic procedure is to first seed the optimization by selecting $k_0$ hyperparameters from $\Lambda$ at random, and then fit the \ac{GP} models for the privacy and utility oracles based on these points.
We then maximize of the \ac{HVPoI} acquisition function to obtain the next query point, which is then added into the dataset.
This is repeated $k$ times until the optimization budget is exhausted.

\vspace*{-0ex}
\begin{algorithm}
\DontPrintSemicolon
\caption{\DPareto{}}\label{alg:main}
\KwIn{hyperparameter set $\Lambda$, privacy and utility oracles $\privor_{\delta}, \utilor_{z}$, anti-ideal point $\antiideal$, number of initial points $k_0$, number of iterations $k$, prior GP}
Initialize dataset $\Dcal \leftarrow \emptyset$\;
\For{$i \in [k_0]$}{
Sample random point $\lambda \in \Lambda$\;
Evaluate oracles $v \leftarrow (\privor_{\delta}(\lambda), 1 - \utilor_{z}(\lambda))$\;
Augment dataset $\Dcal \leftarrow \Dcal \cup \{(\lambda,v)\}$\;
}
\For{$i \in [k]$}{
Fit \acp{GP} to transformed privacy and utility using $\Dcal$\;
Obtain new query point $\lambda$ by optimizing \ac{HVPoI} in \labelcref{eq:hvpoi}
using anti-ideal point $\antiideal$\;
Evaluate oracles $v \leftarrow (\privor_{\delta}(\lambda), 1 - \utilor_{z}(\lambda))$\;
Augment dataset $\Dcal \leftarrow \Dcal \cup \{(\lambda,v)\}$\;
}
\KwRet{Pareto front $\pareto(\{v \;|\; (\lambda,v) \in \Dcal\})$}
\end{algorithm}

\vspace*{-3ex}
\paragraph*{A Note on Output Domains}
The output domains for the privacy and utility oracles may not be well-modeled by a \gp{}, which models outputs on the entire real line.
For instance, the output domain for the privacy oracle is $[0,+\infty]$.
The output domain for the utility oracle depends on the chosen measure of utility.
A common choice of utility oracle for ML tasks is accuracy, which has output domain $[0,1]$.
Thus, neither the privacy nor utility oracles are well-modeled by a \gp{} as-is.
Therefore, in both cases, we transform the outputs so that we are modeling a \gp{} in the transformed space.
For privacy, we use a simple log transform; for accuracy, we use a logit transform $\operatorname{logit}(x) = \log(x) - \log(1 - x)$.
With this, both oracles have transformed output domain $[-\infty,+\infty]$.
Note that it is possible to \textit{learn} the transformation using Warped \acp{GP} \cite{snelson2004warped}.
The advantage there is that the form of both the covariance matrix and the nonlinear transformation are learnt simultaneously under the same probabilistic framework.
However, for simplicity and efficiency we choose to use fixed transformations.

\subsection{Two Illustrative Examples: Revisited} \label{sec:illustrations-revisited}
We revisit the examples discussed in \Cref{sec:illustrations} to concretely illustrate how the components of \DPareto{} work together to effectively learn the privacy--utility Pareto front.

\begin{figure*}[h]
\begin{center}
\includegraphics[clip,width=.77\textwidth]{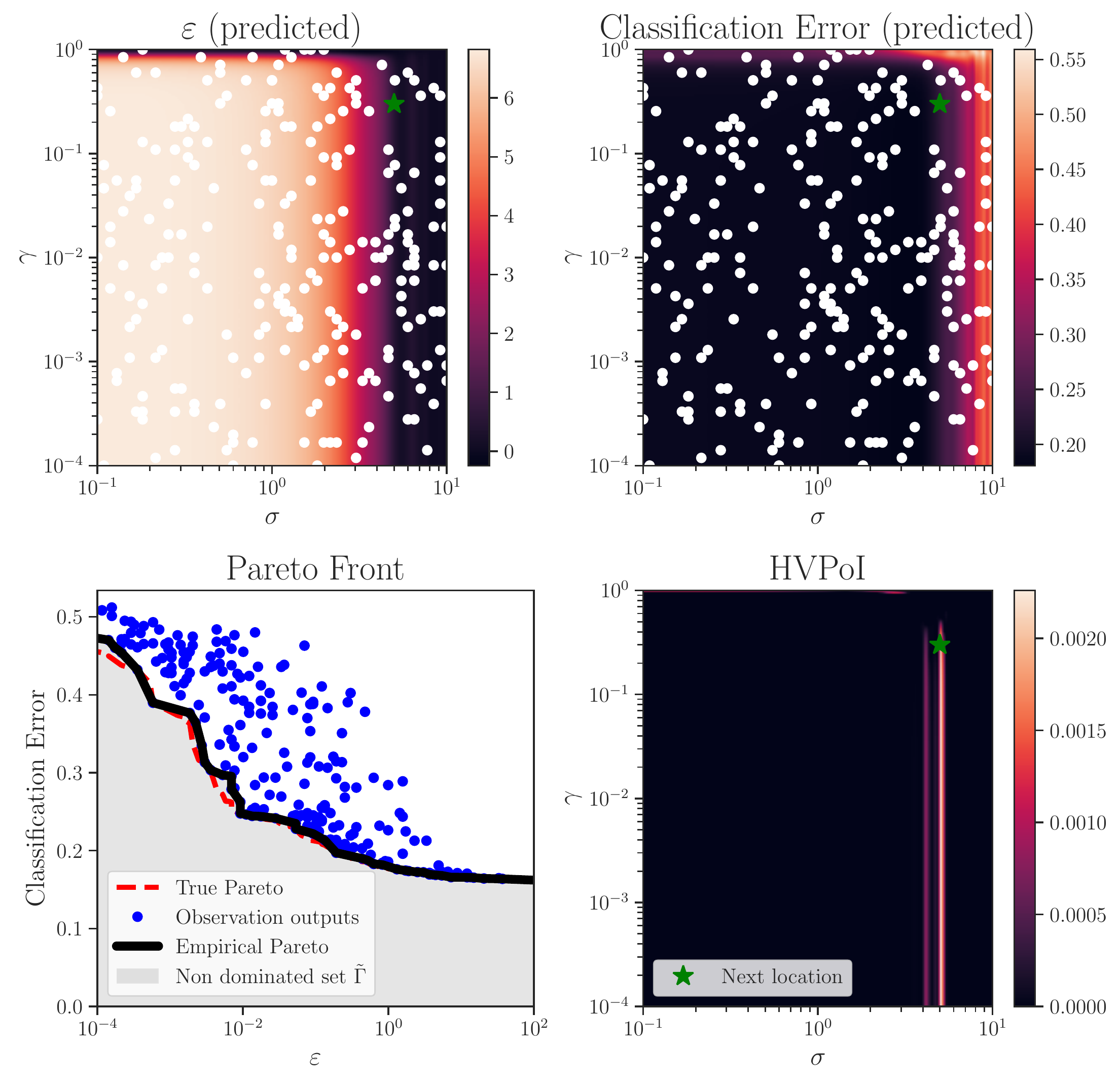}
\end{center}
\vspace*{-3ex}
\caption{\emph{Top:} Mean predictions of the privacy ($\varepsilon$) and the utility (classification error) oracles using their respective \acp{GP} models in the private logistic regression example. The locations of the $k_0 = 250$ sampled points are plotted in white. \emph{Bottom left:} Empirical and true Pareto fronts. \emph{Bottom right:} HVPoI and the selected next location.}
\label{fig:psgd_predictions}
\end{figure*}

\vspace*{-3ex}
\paragraph*{Private Logistic Regression}
For this example, we initialize the \ac{GP} models with $k_0 = 250$ random hyperparameter pairs $(\gamma_i,\sigma_i)$.
$\gamma_i$ takes values in $[10^{-4}, 10^0]$ and $\sigma_i$ takes values in $[10^{-1}, 10^1]$, both sampled uniformly on a logarithmic scale.
The privacy and mean utility of the trained models corresponding to each sample are computed, and \acp{GP} are fit to these values as surrogate models for each oracle.
The predicted means of these surrogate models are shown in the top row of \Cref{fig:psgd_predictions}.
Comparing directly to the oracles' true values in \Cref{fig:psgd_exact}, we observe that the surrogate models have modeled them well in the high $\sigma$ and $\gamma$ regions, but is still learning the low regions.
The bottom-left of \Cref{fig:psgd_predictions} shows the exact Pareto front of the problem, along with the output values of the initial sample and the corresponding empirical Pareto front.
The empirical Pareto front sits almost exactly on the true one, except in the extremely-high privacy region ($\varepsilon < 10^{-2}$) -- this indicates that the selection of random points $(\gamma_i, \sigma_i)$ was already quite good outside of this region.
The goal of \DPareto{} is to select new points in the input domain whose outputs will bring the empirical front closer to the true one.
This is the purpose of the HVPoI function; the bottom-right of \Cref{fig:psgd_predictions} shows the HVPoI function evaluated over all $(\gamma_i, \sigma_i)$ pairs.
The maximizer of this function, marked with a star, is used as the next location to evaluate the oracles.
Note that given the current surrogate models, the HVPoI seems to be making a sensible choice: selecting a point where $\varepsilon$ and classification error are both predicted to have relatively low values, possibly looking to improve the upper-left region of the Pareto front.

\begin{figure*}[h]
\begin{center}
\includegraphics[clip,width=.77\textwidth]{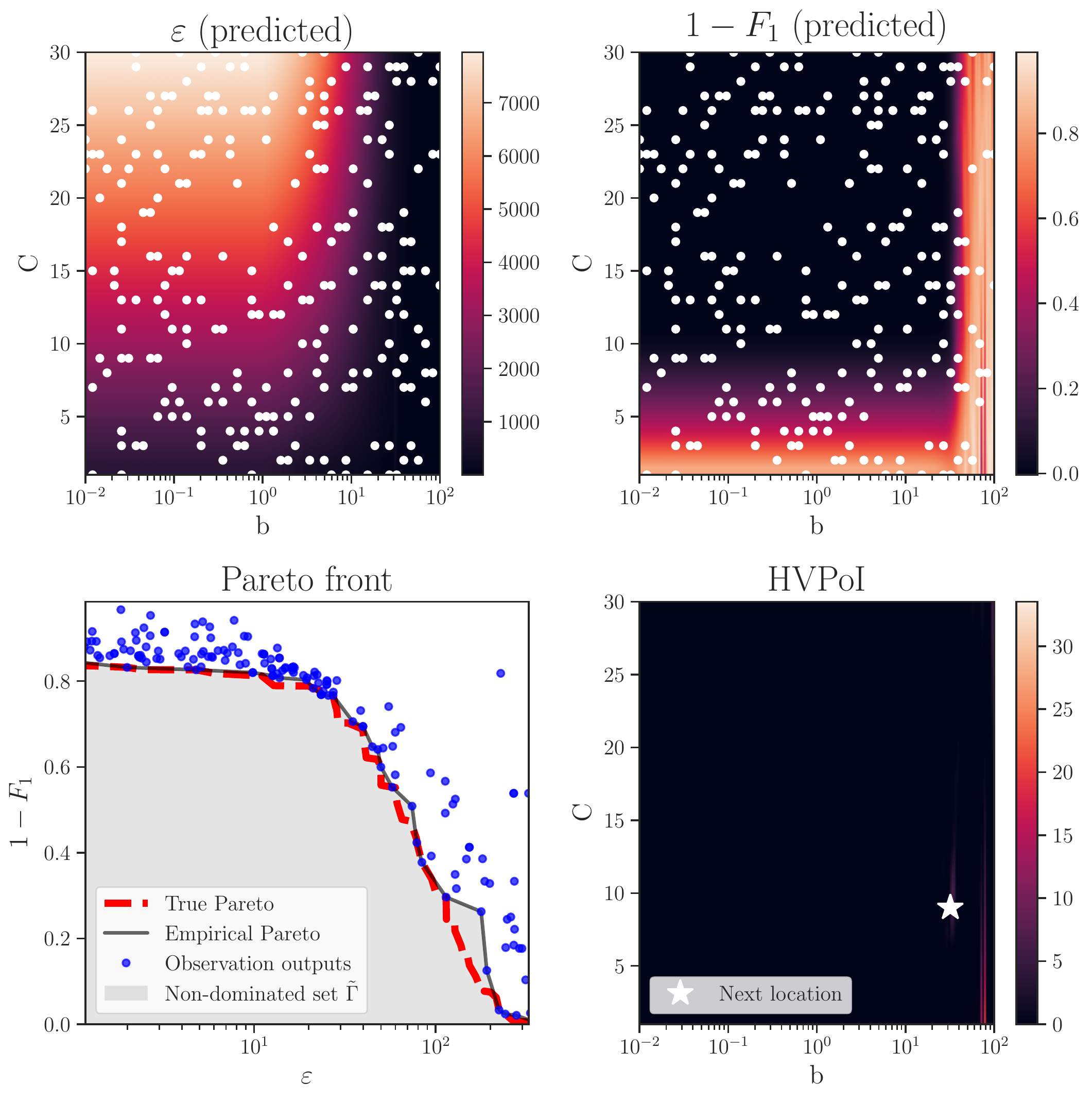}
\end{center}
\vspace*{-3ex}
\caption{\emph{Top:} Mean predictions of the privacy ($\varepsilon$) and the utility ($1 - F_1$) oracles using their respective \acp{GP} models in the sparse vector technique example. The locations of the $k_0 = 250$ sampled points are plotted in white. \emph{Bottom left:} Empirical and true Pareto fronts. \emph{Bottom right:} HVPoI and the selected next location.}\label{fig:sparse_vector_predicted}
\vspace*{-1ex}
\end{figure*}

\vspace*{-3ex}
\paragraph*{Sparse Vector Technique}
For this example, we initialize the \ac{GP} models with $k_0 = 250$ random hyperparameter pairs $(C_i,b_i)$.
The $C_i$ values are sampled uniformly in the interval $[1,30]$, and the $b_i$ values are sampled uniformly in the interval $[10^{-2}, 10^2]$ on a logarithmic scale.
The privacy and utility values are computed for each of the samples, and \acp{GP} are fit to these values as surrogate models for each oracle.
The predicted means of these surrogate models are shown in the top row of \Cref{fig:sparse_vector_predicted}.
We observe that both surrogate models have modeled their oracles reasonably well, comparing directly to the oracles' true values in \Cref{fig:sparse_vector_exact}.
The bottom-left of \Cref{fig:sparse_vector_predicted} shows the exact Pareto front of the problem, along with the output values of the initial sample and the corresponding empirical Pareto front.
The empirical Pareto front sits close to the true one, which indicates that the selection of points $(C_i,b_i)$ is already quite good.
The HVPoI function is used by \DPareto{} to select new points in the input domain whose outputs will bring the empirical front closer to the true one.
The bottom-right of \Cref{fig:sparse_vector_predicted} shows this function evaluated over all $(C_i,b_i)$ pairs.
The maximizer of this function, marked with a star, is used as the next location to evaluate the oracles.
Note that given the current surrogate models, the HVPoI appears to be making a sensible choice: selecting a point where $\varepsilon$ is predicted to have a medium value and $1 - F_1$ is predicted to have a low value, possibly looking to improve the gap in the lower-right corner of the Pareto front.

\section{Experiments}\label{sec:experiments}

In this section, we provide experimental evaluations of \DPareto{} on a number of ML tasks.
Unlike the illustrations previously discussed in \Cref{sec:illustrations,sec:illustrations-revisited}, it is computationally infeasible to compute exact Pareto fronts for these tasks.
This highlights the advantage of using \DPareto{} over random and grid search baselines, showcasing its versatility on a variety of models, datasets, and optimizers. See Appendix \ref{sec:implementation} for implementation details.

\subsection{Experimental Setup}\label{sec:exp-setup}
In all our experiments we used $v^{\dagger} = (10, 1)$ as the anti-ideal point in \DPareto, encoding our interest in a Pareto front which captures a practical privacy range (i.e., $\varepsilon \le 10$) across all possible utility values (since classification error can never exceed $1$).

\vspace*{-3ex}
\paragraph*{Optimization Domains and Sampling Distributions} \Cref{tab:domains} gives the optimization domain $\Lambda$ for each of the different experiments, which all point-selection approaches (i.e., \DPareto, random sampling, and grid search) operate within.
Random sampling distributions for experiments on both \MNIST{} and \Adult{} datasets were carefully constructed in order to generate as favorable results for random sampling as possible.
These distributions are precisely detailed in \Cref{sec:rsdists}, and were constructed by reviewing literature (namely Abadi et al.~\cite{abadi2016deep} and McMahan et al.~\cite{brendan2018learning}) in addition to the authors' experience from training these differentially private models.
The Pareto fronts generated from these constructed distributions were significantly better than those yielded by the na{\"i}ve strategy of sampling from the uniform distribution, justifying the choice of these distributions.

\begin{table*}[ht]
\centering
\adjustbox{max width=\linewidth}{%
\begin{tabular}{@{}lllllll@{}}
\toprule
\textbf{Algorithm} & \textbf{Dataset} & \textbf{Epochs} ($T$)       & \textbf{Lot Size} ($m$)       & \textbf{Learning Rate} ($\eta$)   & \textbf{Noise Variance}  ($\sigma^2$) & \textbf{Clipping Norm} ($L$) \\ \midrule
LogReg+SGD         & \Adult{}            & $\left[1, 64\right]$  & $\left[8, 512\right]$   & $\left[\num{5e-4}, \num{5e-2}\right]$ & $\left[0.1, 16\right]$           & $\left[0.1, 4\right]$            \\
LogReg+Adam        & \Adult{}            & $\left[1, 64\right]$  & $\left[8, 512\right]$   & $\left[\num{5e-4}, \num{5e-2}\right]$ & $\left[0.1, 16\right]$           & $\left[0.1, 4\right]$            \\
SVM+SGD         & \Adult{}            & $\left[1, 64\right]$  & $\left[8, 512\right]$   & $\left[\num{5e-4}, \num{5e-2}\right]$ & $\left[0.1, 16\right]$           & $\left[0.1, 4\right]$            \\
MLP1+SGD           & \MNIST{}            & $\left[1, 400\right]$ & $\left[16, 4000\right]$ & $\left[\num{1e-3}, \num{5e-1}\right]$ & $\left[0.1, 16\right]$           & $\left[0.1, 12\right]$           \\
MLP2+SGD           & \MNIST{}            & $\left[1, 400\right]$ & $\left[16, 4000\right]$ & $\left[\num{1e-3}, \num{5e-1}\right]$ & $\left[0.1, 16\right]$           & $\left[0.1, 12\right]$           \\ \bottomrule
\end{tabular}
}
\caption{Optimization domains used in each of the experimental settings.}
\label{tab:domains}
\end{table*}

\vspace*{-2ex}
\paragraph*{Datasets} We tackle two classic problems: multiclass classification of handwritten digits with the \MNIST{} dataset, and binary classification of income with the \Adult{} dataset.
\MNIST{} \cite{lecun1998gradient} is composed of $28 \times 28$ gray-scale images, each representing a single digit $0$-$9$. It has $60$k ($10$k) images in the training (test) set.
\Adult{} \cite{kohavi1996scaling} is composed of $123$ binary demographic features on various people, with the task of predicting whether income > \$$50$k. It has $40$k ($1.6$k) points in the training (test) set.

\vspace*{-2ex}
\paragraph*{Models} For \Adult{} dataset, we consider \ac{LogReg} and linear \acp{SVM}, and explore the effect of the choice of model and optimization algorithm (SGD vs.\ Adam), using the differentially private versions of these algorithms outlined in \Cref{sec:exp-priv-algos}. 
For \MNIST{}, we fix the optimization algorithm as SGD, but use a more expressive \ac{MLP} model and explore the choice of network architectures. The first (MLP1) has a single hidden layer with $1000$ neurons, which is the same as used by Abadi et al.~\cite{abadi2016deep} but without DP-PCA dimensionality reduction.
The second (MLP2) has two hidden layers with $128$ and $64$ units.
In both cases we use ReLU activations.

\subsection{Privatized Optimization Algorithms}\label{sec:exp-priv-algos}
Experiments are performed with privatized variants of two popular optimization algorithms -- \ac{SGD} \cite{bottou2010large} and Adam \cite{kingma2015adam}
-- although our framework can easily accommodate other privatized algorithms when a privacy oracle is available. \Acf{SGD} is a simplification of gradient descent, where on each iteration instead of computing the gradient for the entire dataset, it is instead estimated on the basis of a single example (or small batch of examples) picked uniformly at random (without replacement) \cite{bottou2010large}. Adam \cite{kingma2015adam} is a first-order gradient-based optimization algorithm for stochastic objective functions, based on adaptive estimates of lower-order moments. 

As a privatized version of SGD, we use a mini-batched implementation with clipped gradients and Gaussian noise, detailed in \Cref{alg:dpsgd}.
This algorithm is similar to that of Abadi et al.'s~\cite{abadi2016deep}, but differs in two ways.
First, it utilizes sampling without replacement to generate fixed-size mini-batches, rather than using Poisson sampling with a fixed probability which generates variable-sized mini-batches.
Using fixed-size mini-matches is a is a more natural approach, which more-closely aligns with standard practice in non-private ML.
Second, as the privacy oracle we use the moments accountant implementation of Wang et al.~\cite{wang-balle-kasiviswanathan18}, which supports sampling without replacement.
In \Cref{alg:dpsgd}, the function $\clip_{L}(v)$ acts as the identity if $\norm{v}_2 \leq L$, and otherwise returns $(L / \norm{v}_2) v$. This clipping operation ensures that $\norm{\clip_{L}(v)}_2 \leq L$ so that the $\ell_2$-sensitivity of any gradient to a change in one datapoint in $z$ is always bounded by $L/m$.

\begin{algorithm}%
\DontPrintSemicolon
\SetKwInput{KwHP}{Hyperparameters}
\caption{Differentially Private SGD}\label{alg:dpsgd}
\KwIn{dataset $z = (z_1, \ldots, z_n)$}
\KwHP{learning rate $\eta$, mini-batch size $m$, number of epochs $T$, noise variance $\sigma^2$, clipping norm $L$}
Initialize $w \leftarrow 0$\;
\For{$t \in [T]$}{
\For{$k \in [n/m]$}{
Sample $S \subset [n]$ with $|S| = m$ uniformly at random\;
Let $g \leftarrow \frac{1}{m} \sum_{j \in S} \clip_{L}(\nabla \ell(z_j, w)) + \frac{2 L}{m} \cN(0,\sigma^2 I)$\;
Update $w \leftarrow w - \eta g$\;
}
}
\KwRet{$w$}
\end{algorithm}

Our privatized version of Adam is given in \Cref{alg:dpadam}, which uses the same gradient perturbation technique as stochastic gradient descent.
Here the notation $g^{\odot 2}$ denotes the vector obtained by squaring each coordinate of $g$.
Adam uses three numerical constants that are not present in SGD ($\kappa$, $\beta_1$ and $\beta_2$). To simplify our experiments, we fixed those constants to the defaults suggested in Kingma et al.~\cite{kingma2015adam}.

\vspace*{-2ex}
\begin{algorithm}[h]%
\DontPrintSemicolon
\SetKwInput{KwHP}{Hyperparameters}
\caption{Differentially Private Adam}\label{alg:dpadam}
\KwIn{dataset $z = (z_1, \ldots, z_n)$}
\KwHP{learning rate $\eta$, mini-batch size $m$, number of epochs $T$, noise variance $\sigma^2$, clipping norm $L$}
Fix $\kappa \leftarrow 10^{-8}$, $\beta_1 \leftarrow 0.9$, $\beta_2 \leftarrow 0.999$\;
Initialize $w \leftarrow 0$, $\mu \leftarrow 0$, $\nu \leftarrow 0$, $i \leftarrow 0$\;
\For{$t \in [T]$}{
\For{$k \in [n/m]$}{
Sample $S \subset [n]$ with $|S| = m$ uniformly at random\;
Let $g \leftarrow \frac{1}{m} \sum_{j \in S} \clip_{L}(\nabla \ell(z_j, w)) + \frac{2 L}{m} \cN(0,\sigma^2 I)$\;
Update $\mu \leftarrow \beta_1 \mu + (1-\beta_1) g$, $\nu \leftarrow \beta_2 \nu + (1-\beta_2) g^{\odot 2}$, $i \leftarrow i + 1$\;
De-bias $\hat{\mu} \leftarrow \mu / (1 - \beta_1^i)$, $\hat{\nu} \leftarrow \nu / (1 - \beta_2^i)$\;
Update $w \leftarrow w - \eta \hat{\mu} / (\sqrt{\hat{\nu}} + \kappa)$\;
}
}
\KwRet{$w$}
\end{algorithm}

\subsection{Experimental Results}\label{sec:exp-results}

\begin{figure*}[t]
\begin{center}
\includegraphics[clip,width=0.8\textwidth]{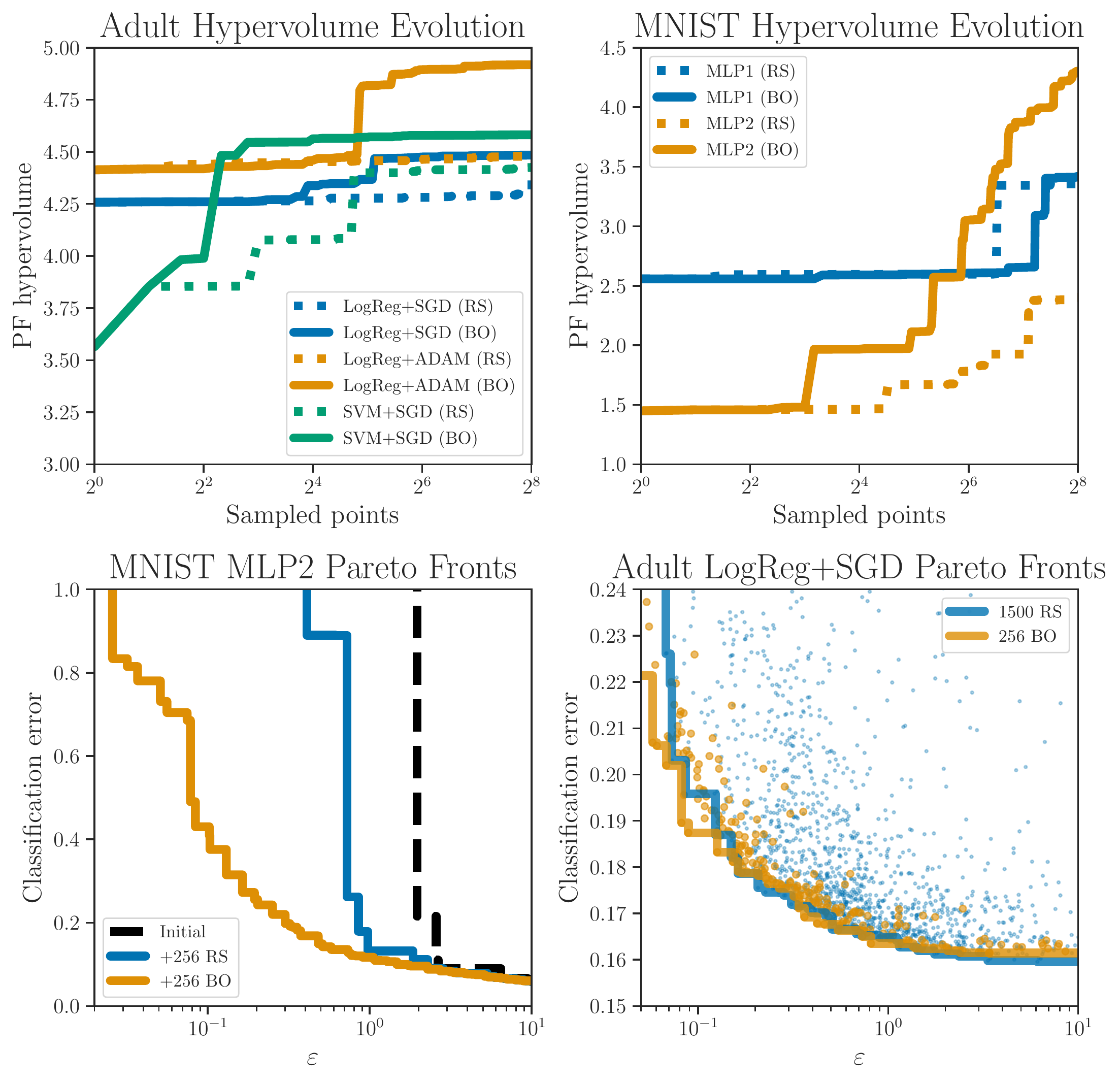}
\end{center}
\vspace*{-3ex}
\caption{\emph{Top:} Hypervolumes of the Pareto fronts computed by the various models, optimizers, and architectures on the \Adult{} and \MNIST{} datasets (respectively) by both \DPareto{} and random sampling. \emph{Bottom left:} Pareto fronts learned for MLP2 architecture on the \MNIST{} dataset with \DPareto{} and random sampling, including the shared points they were both initialized with. \emph{Bottom right:} \Adult{} dataset \DPareto{} sampled points and its Pareto front compared with the larger set of random sampling points and its Pareto front.}
\label{fig:bo_vs_rs}
\end{figure*}

\paragraph*{\DPareto{} vs.\ Random Sampling}
A primary purpose of these experiments is to highlight the efficacy of \DPareto{} at estimating an algorithm's Pareto front.
As discussed above, the hypervolume is a popular measure for quantifying the quality of a Pareto front.
We compare \DPareto{} to the traditional approach of random sampling by computing the hypervolumes of Pareto fronts generated by each method.

In \Cref{fig:bo_vs_rs} the first two plots show, for a variety of models, how the hypervolume of the Pareto front expands as new points are sampled.
In nearly every experiment, the \DPareto{} approach yields a greater hypervolume than the random sampling analog -- a direct indicator that \DPareto{} has better characterized the Pareto front.
This can be seen by examining the bottom left plot of the figure, which directly shows a Pareto front of the MLP2 model with both sampling methods.
Specifically, while the random sampling method only marginally improved over its initially seeded points, \DPareto{} was able to thoroughly explore the high-privacy regime (i.e.\ small $\varepsilon$).
The bottom right plot of the figure compares the \DPareto{} approach with 256 sampled points against the random sampling approach with significantly more sampled points, $1500$.
While both approaches yield similar Pareto fronts, the efficiency of \DPareto{} is particularly highlighted by the points that are \emph{not} on the front: nearly all the points chosen by \DPareto{} are close to the actual front, whereas many points chosen by random sampling are nowhere near it. 

To quantify the differences between random sampling and \DPareto{} for the \Adult{} dataset, we split the $5000$ random samples into $19$ parts of size $256$ to match the number of \ac{BO} points, and computed hypervolume differences between the resultant Pareto fronts under the mild assumption that \DPareto{} is deterministic\footnote{While not strictly true, since \ac{BO} is seeded with a random set of points, running repetitions would have been an extremely costly exercise with results expected to be nearly identical.}.
We then computed the two-sided confidence intervals for these differences, shown in \Cref{tab:statistical_tests}.
We also computed the t-statistic for these differences being zero, which were all highly significant $(p<0.001)$. This demonstrates that the observed differences between Pareto fronts are in fact statistically significant.
We did not have enough random samples to run statistical tests for \MNIST{}, however the differences are visually even clearer in this case.

\begin{table}[!htbp]
\centering
\begin{tabular}{@{}lll@{}}
\toprule
Algorithm+Optimizer & Mean Difference & \textbf{95\% C.I.} \\ \midrule
LogReg+SGD	& 0.158	& $(0.053, 0.264)^*$    \\
LogReg+ADAM	& 0.439	& $(0.272, 0.607)^*$    \\
SVM+SGD	& 0.282	& $(0.161, 0.402)^*$   \\
\end{tabular}
\caption{Mean hypervolume differences between BO and $19$ random repetitions of $256$ iterations of random sampling. Two-sided $95\%$ confidence intervals (C.I.) for these differences, as well as t-tests for the mean, are included. Asterisks indicate significance at the $p<0.001$ level.}
\label{tab:statistical_tests}
\end{table}

\begin{figure}[htp]
\begin{center}
\includegraphics[clip,width=\textwidth]{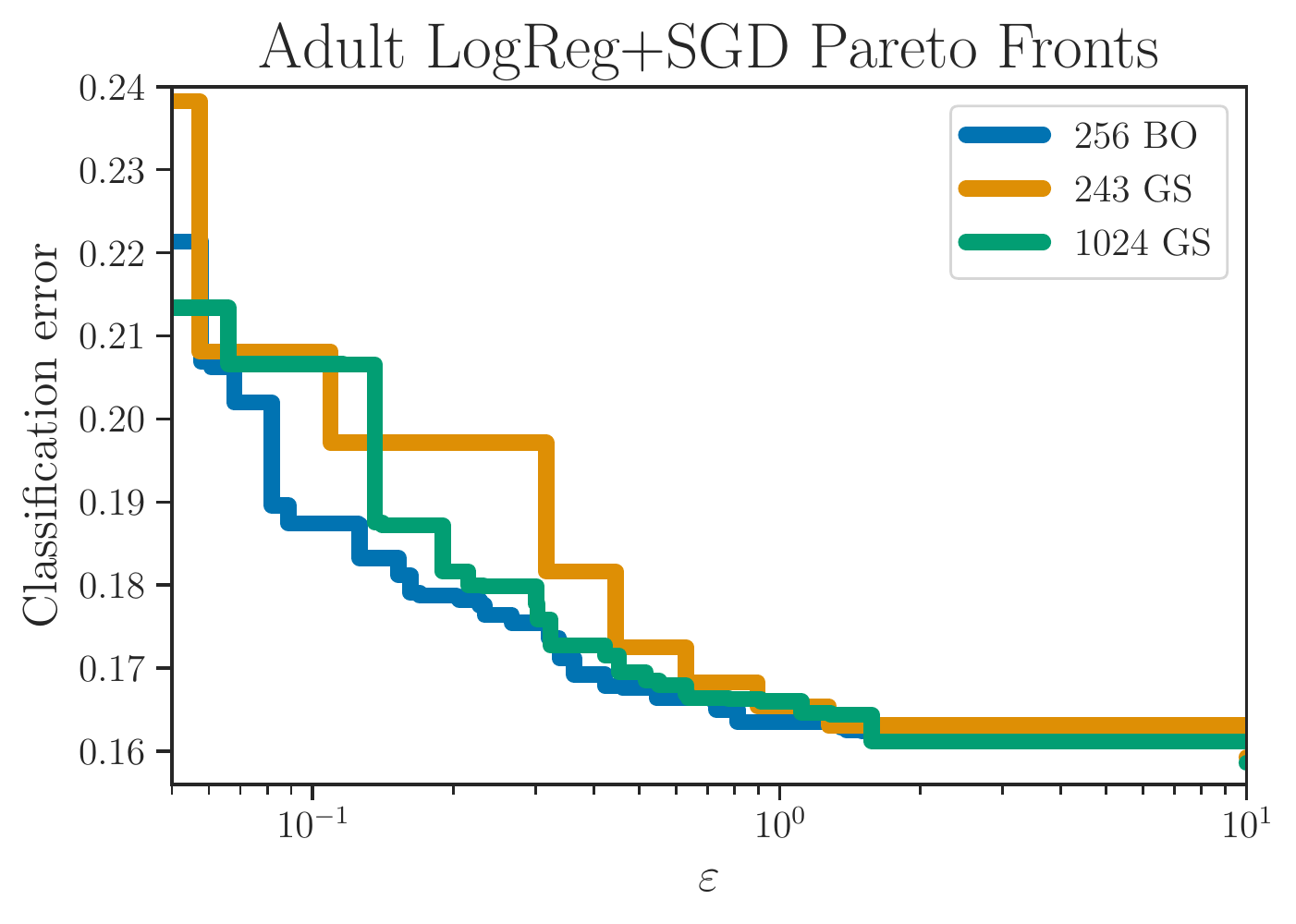}
\end{center}
\vspace*{-3ex}
\caption{Grid search experiment results compared with the Bayesian optimization approach used in \DPareto.}
\label{fig:grid_vs_bayesopt}
\end{figure}

\vspace*{-2ex}
\paragraph*{\DPareto{} vs.\ Grid search}
For completeness we also ran experiments using grid search with two different grid sizes, both of which performed significantly worse than \DPareto{}.
For these experiments, we have defined parameter ranges as the limiting parameter values from our random sampling experiment setup (see  \Cref{tab:adult-sampling-dist}).
We evaluated a grid size of $3$, which corresponds to $243$ total points (approximately the same amount of points as \DPareto{} uses), and grid size $4$, which corresponds to $1024$ points ($4$ times more than were used for \DPareto).
As can be seen in \Cref{fig:grid_vs_bayesopt}, \DPareto{} outperforms grid search even when significantly more grid points are evaluated.

\begin{figure*}[htp]
\begin{center}
\includegraphics[clip,width=\textwidth]{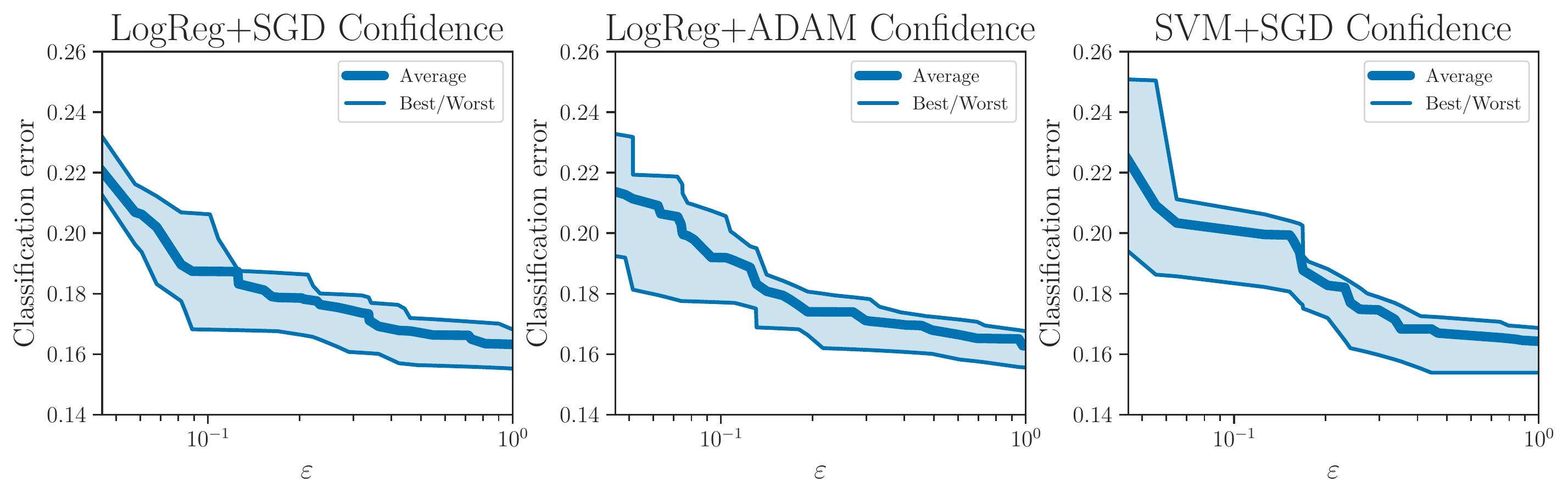}
\end{center}
\vspace*{-4ex}
\caption{Variability of estimated Pareto fronts across models and optimizers for \Adult{}.}
\label{fig:pf_confidence}
\vspace*{-2ex}
\end{figure*}

\vspace*{-6ex}
\paragraph*{Variability of Pareto Front}
\DPareto{} also allows us to gather information about the potential variability of the recovered Pareto front.
In order to do that, recall that in our experiments we implemented the utility oracle by repeatedly running algorithm $A_{\lambda}$ with a fixed choice of hyperparameters, and then reported the average utility across runs.
Using these same runs we can also take the best and worst utilities observed for each choice of hyperparameters.
\Cref{fig:pf_confidence} displays the Pareto fronts recovered from considering the best and worst runs in addition to the Pareto front obtained from the average over runs.
In general we observe higher variability in utility on the high privacy regime (i.e.\ small $\varepsilon$), which is to be expected since more privacy is achieved by increasing the variance of the noise added to the computation.
These type of plots can be useful to decision-makers who want to get an idea of what variability can be expected in practice from a particular choice of hyperparameters.

\begin{figure*}[h]
\begin{center}
\includegraphics[clip,width=.75\linewidth]{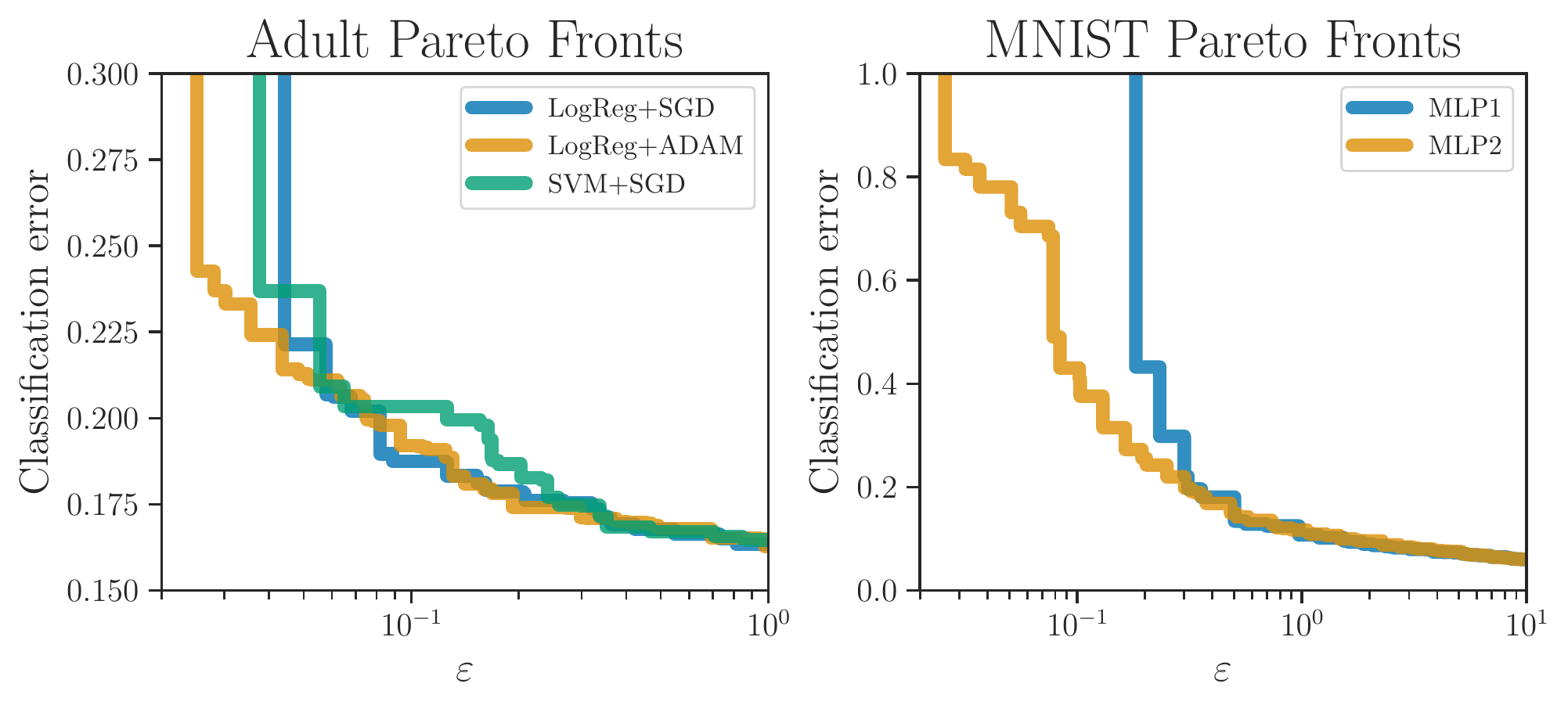}
\end{center}
\vspace*{-4ex}
\caption{\emph{Left:} Pareto fronts for combinations of models and optimizers on the \Adult{} dataset. \emph{Right:} Pareto fronts for different \ac{MLP} architectures on the \MNIST{} dataset.}
\label{fig:pf_comparison}
\end{figure*}

\vspace*{-2ex}
\paragraph*{\DPareto{}'s Versatility}
The other main purpose of these experiments is to demonstrate the versatility of \DPareto{} by comparing multiple approaches to the same problem.
In \Cref{fig:pf_comparison}, the left plot shows Pareto fronts of the \Adult{} dataset for multiple optimizers (SGD and Adam) as well as multiple models (\ac{LogReg} and \ac{SVM}), and the right plot shows Pareto fronts of the \MNIST{} dataset for different architectures (MLP1 and MLP2).
With this, we can see that on the \Adult{} dataset, the \ac{LogReg} model optimized using Adam was nearly always better than the other model/optimizer combinations.
We can also see that on the \MNIST{} dataset, while both architectures performed similarly in the low-privacy regime, the MLP2 architecture significantly outperformed the MLP1 architecture in the high-privacy regime.
With \DPareto{}, analysts and practitioners can efficiently create these types of Pareto fronts and use them to perform privacy--utility trade-off comparisons.

\vspace*{-2ex}
\paragraph*{Computational Overhead of \DPareto{}}
Although it is clear that \DPareto{} more-efficiently produces high-quality Pareto fronts relative to random sampling and grid search, we must examine the computational cost it incurs.
Namely, we are interested in the running time of \DPareto{}, excluding the model training time.
Therefore, for the BO experiments on both datasets, we measured the time it took for \DPareto{} to: 1) initialize the GP models with the $16$ seed points, plus 2)  iteratively propose the subsequent $256$ hyperparameters and incorporate their corresponding privacy and utility results.

For both the \Adult{} and \MNIST{} datasets, despite the difference in hyperparameter domains as well as the privacy and utility values that were observed, \DPareto{}'s overhead remained fairly consistent at approximately $45$ seconds of total wall-clock time.
This represents a negligible fraction of the total Pareto front computation time for either dataset; specifically, less than $0.1\%$ of the total time for the \Adult{} Pareto fronts, and less than $0.01\%$ for the \MNIST{} Pareto fronts.
Thus, we conclude that \DPareto{}'s negligible overhead is more than offset by its improved Pareto fronts.

We remark here that although the overhead is negligible, \DPareto{} does have a shortcoming relative to traditional methods: it is an inherently sequential process which cannot be easily parallelized.
On the other hand, random search and grid search can be trivially parallelized to an arbitrarily high degree bounded only by one's compute resources.
Improving upon this facet of \DPareto{} is beyond the scope of this work, however we briefly discuss it as a possible extension in the following section.

\section{Extensions} \label{sec:extensions}

There are several extensions and open problems whose solutions would enhance \DPareto{}'s usefulness for practical applications.

The first open problem is on the privacy side. 
As designed, \DPareto{} is a system to \emph{non-privately} estimate the Pareto front of differentially private algorithms.
One challenging open problem is how to tightly characterize the privacy guarantee of the estimated Pareto front itself.
This involves analyzing the privacy guarantees for both the training and test data sets.
Na\"ively applying \ac{DP} composition theorems immediately provides conservative bounds on the privacy for both sets (assuming a small amount of noise is added to the output of the utility oracle).
This follows from observing that individual points evaluated by \DPareto{} enjoy the \ac{DP} guarantees computed by the privacy oracle, and the rest of the algorithm only involves post-processing and adaptive composition.
However, these bounds would be prohibitively large for practical use;
we expect a more advanced analysis to yield significantly tighter guarantees since for each point we are only releasing its utility and not the complete trained model.
For a decision-maker, these would provide an end-to-end privacy guarantee for the entire process, and allow the Pareto front to be made publicly available.

The other open problem is on the \ac{BO} side.
Recall that the estimated Pareto front contains only the privacy--utility values of the trained models, along with their corresponding hyperparameters.
In practice, a decision-maker may be interested in finding a hyperparameter setting that induces a particular point on the estimated Pareto front but which was not previously tested by \DPareto{}.
The open problem here is how to design a computationally efficient method to extract this information from \DPareto{}'s \acp{GP}.

On the theoretical side, it would also be interesting to obtain bounds on the speed of convergence of \DPareto{} to the optimal Pareto front.
Results of this kind are known for \emph{single-objective} \ac{BO} under smoothness assumptions on the objective function (see, e.g., \cite{bull2011convergence}).
Much less is known in the \emph{multi-objective} case, especially in a setting like ours that involves continuous hyper-parameters and noisy observations (from the utility oracle).
Nonetheless, the smoothness we observe on the privacy and utility oracles as a function of the hyperparameters, and our empirical evaluation against grid and random sampling, suggests similar guarantees could hold for \DPareto{}.

For applications of \DPareto{} to concrete problems, there are several interesting extensions.
We focused on supervised learning, but the method could also be applied to e.g.\ stochastic variational inference on probabilistic models, as long as a utility function (e.g. held-out perplexity) is available.
\DPareto{} currently uses independent \acp{GP} with fixed transformations, but an interesting extension would be to use warped multi-output \acp{GP}.
It may be of interest to optimize over additional criteria, such as model size, training time, or a fairness measure.
If the sequential nature of \ac{BO} is prohibitively slow for the problem at hand, then adapting the recent advances in batch multi-objective \ac{BO} to \DPareto{} would enable parallel evaluation of several candidate models \cite{gaudrie2020targeting, lin2018batched, wang2019balancing}.
Finally, while we explored the effect of changing the model (logistic regression vs.\ \ac{SVM}) and the optimizer (SGD vs.\ Adam) on the privacy--utility trade-off, it would be interesting to optimize over these choices as well.

\section{Related Work}\label{sec:related}

While this work is the first to examine the privacy--utility trade-off of differentially private algorithms using multi-objective optimization and Pareto fronts, efficiently computing Pareto fronts without regards to privacy is an active area of research in fields relating to multi-objective optimization.
\DPareto's point-selection process aligns with Couckuyt et al.~\cite{Couckuyt:2014}, but other approaches may provide promising alternatives for improving \DPareto{}.
For example, Zuluaga et al.~\cite{zuluaga2016varepsilon} propose an acquisition criteria that focuses on uniform approximation of the Pareto front instead of a hyper-volume based criteria.
Note that their technique does not apply out-of-the-box to the problems we consider in our experiments since it assumes a discrete hyper-parameter space.

The threat model and outputs of the \DPareto{} algorithm are closely aligned with the methodology used by the U.S.\ Census Bureau to choose the privacy parameter $\varepsilon$ for their deployment of \ac{DP} to release data from the upcoming 2020 census.
In particular, the bureau is combining a graphical approach to represent the privacy--utility trade-off for their application \cite{garfinkel2018issues} together with economic theory to pick a particular point to balance the trade-off \cite{abowd2018economic}.
Their graphical approach works with Pareto fronts identical to the ones computed by our algorithm, which they construct using data from previous censuses \cite{fall-csac}.
We are not aware of the specifics of their hyperparameter tuning, but, from what has transpired, it would seem that the gross of hyperparameters in their algorithms is related to the post-processing step and therefore only affects utility\footnote{Or, in the case of invariant forcing, privacy effects not quantifiable within standard \ac{DP} theory.}.

Several aspects of this paper are related to recent work in single-objective optimization.
For non-private single-objective optimization, there is an abundance of recent work in machine learning on hyperparameter selection, using BO \cite{klein2017fast,golovin2017google} or other methods \cite{li2017hyperband} to maximize a model's utility.
Recently, several related questions at the intersection of machine learning and differential privacy have emerged regarding hyperparameter selection and utility.

One such question explicitly asks how to perform the hyperparameter-tuning process in a privacy-preserving way.
Kusner et al.~\cite{kusner2015differentially} and subsequently Smith et al.~\cite{smith2018differentially} use BO to find near-optimal hyperparameter settings for a given model while preserving the privacy of the data during the utility evaluation stage.
Aside from the single-objective focus of this setting, our case is significantly different in that we are primarily interested in \emph{training} the models with differential privacy, not in protecting the privacy of the data used to evaluate an already-trained model.

Another question asks how to choose utility-maximizing hyperparameters when privately training models.
When privacy is independent of the hyperparameters, this reduces to the non-private hyperparameter optimization task. However, two variants of this question do not have this trivial reduction.
The first variant inverts the stated objective to study the problem of maximizing privacy given constraints on the final utility~\cite{ligett2017accuracy,ge2019apex}.
The second variant, closely aligning with this paper's setting, studies the problem of choosing utility-maximizing, but privacy-dependent, hyperparameters.
This is particularly challenging, since the privacy's dependence on the hyperparameters may be non-analytical and computationally expensive to determine.
Approaches to this variant have been studied~\cite{DBLP:journals/corr/abs-1812-06210,van2018practical}, however the proposed strategies are 1) based on heuristics, 2) only applicable to the differentially private SGD problem, and 3) do not provide a computationally efficient way to find the Pareto optimal points for the privacy--utility trade-off of a given model.
Wu et al.~\cite{wu2017bolt} provide a practical analysis-backed approach to privately training utility-maximizing models (again, for the case of SGD with a fixed privacy constraint), but hyperparameter optimization is na{\"i}vely performed using grid-search.
By contrast, this paper provides a computationally efficient way to \emph{directly} search for Pareto optimal points for the privacy--utility trade-off of arbitrary hyperparameterized algorithms.

The final related question revolves around the differentially private ``selection'' or ``maximization'' problem \cite{chaudhuri2014large}, which asks: how can an item be chosen (from a predefined universe) to maximize a data-dependent function while still protecting the privacy of that data?
Here, Liu et al.~\cite{liu2019private} recently provided a way to choose hyperparameters that approximately maximize the utility of a given differentially private model in a way that protects the privacy of both the training and test data sets.
However, this only optimizes utility with fixed privacy -- it does not address our problem of directly optimizing for the selection of hyperparameters that generate privacy--utility points which fall on the Pareto front.

Recent work on data-driven algorithm configuration 
has considered the problem of tuning the hyperparameters of combinatorial optimization algorithms while maintaining \ac{DP} \cite{balcan2018dispersion}.
The setting considered in \cite{balcan2018dispersion} assumes there is an underlying distribution of problem instances, and a sample from this distribution is used to select hyperparameters that will have good computational performance on future problem instances sampled from the same distribution.
In this case, the authors consider a threat model where the whole sample of problem instances used to tune the algorithm needs to be protected.
A similar problem of data-driven algorithm selection has been considered, where the problem is to choose the best algorithm to accomplish a given task while maintaining the privacy of the data used~\cite{kotsogiannis2017pythia}.
For both, only the utility objective is being optimized, assuming a fixed constraint on the privacy.

\section{Conclusion}

In this paper, we characterized the privacy--utility trade-off of a hyperparameterized algorithm as a Pareto front learning problem.%
We then introduced \DPareto{}, a method to empirically learn a differentially private algorithm's Pareto front.
\DPareto{} uses \acf{BO}, a state-of-the-art method for \acl{HPO}, to efficiently form the Pareto front by simultaneously optimizing for both privacy and utility.
We evaluated \DPareto{} across various datasets, models, and differentially private optimization algorithms, demonstrating its efficiency and versatility.
Further, we showed that \ac{BO} allows us to construct useful visualizations to aid the decision making process.

\subsection*{Acknowledgments}

This work was funded by Amazon Research Cambridge.

\bibliographystyle{plain}
\bibliography{paper}   %

\appendix

\normalsize
\section{Implementation Details}\label{sec:implementation}
Hyperparameter optimization was done using the GPFlowOpt library \cite{GPflowOpt2017} which offers \gp{}-based Bayesian optimization.
In particular, we used the provided \ac{GP} regression functionality with a Matérn 5/2 kernel, as well as the provided \ac{HVPoI} acquisition function.
Machine learning models used in the paper are implemented with Apache MXNet \cite{chen2015mxnet}. We have made use of the high-level Gluon API whenever possible. However, the privacy accountant implementation that we used (see \cite{wang-balle-kasiviswanathan18}) required low-level changes to the definitions of the models. In order to keep the continuous MXNet execution graph to ensure a fast evaluation of the model, we reverted to the pure MXNet model definitions. Even though this approach requires much more effort to implement the models themselves, it allows for more fine-grained control of how the model is executed, as well as provides a natural way of implementing privacy accounting.

\section{Privacy Proof for Sparse Vector Technique}\label{sec:svt-proof}
Here we provide a proof of the privacy bound for \Cref{alg:SVT} used to implement the privacy oracle $\privor_0$.
The proof is based on observing that our \Cref{alg:SVT} is just a simple re-parametrization of \cite[Alg. 7]{lyu2017understanding} where some of the parameters have been fixed up-front. For concreteness, we reproduce \cite[Alg. 7]{lyu2017understanding} as \Cref{alg:SVTlyu} below.
The result then follows from a direct application of \cite[Thm. 4]{lyu2017understanding}, which shows that \Cref{alg:SVTlyu} is $(\varepsilon_1 + \varepsilon_2, 0)$-DP.

\begin{algorithm}[h]%
\DontPrintSemicolon
\SetKwInput{KwHP}{Hyperparameters}
\caption{Sparse Vector Technique (\cite[Alg. 7]{lyu2017understanding} with $\varepsilon_3 = 0$)}\label{alg:SVTlyu}
\KwIn{dataset $z$, queries $q_1, \ldots, q_m$, sensitivity $\Delta$}
\KwHP{bound $C$, thresholds $T_1, \ldots, T_m$, privacy parameters $\varepsilon_1$, $\varepsilon_2$}
$c \leftarrow 0$, $w \leftarrow (\bot,\ldots,\bot) \in \{\bot,\top\}^m$\;
$\rho \leftarrow \Lap(\Delta / \varepsilon_1)$\;
\For{$i \in [m]$}{
$\nu \leftarrow \Lap(2 C \Delta/ \varepsilon_2)$\;
\If{$q_i(z) + \nu \geq T_i + \rho$}{
$w_i \leftarrow \top$, $c \leftarrow c + 1$\;
\lIf{$c \geq C$}{\KwRet{$w$}	}
}
}
\KwRet{$w$}
\end{algorithm}

Comparing \Cref{alg:SVTlyu} with the sparse vector technique in \Cref{alg:SVT}, we see that they are virtually the same algorithms, where we have fixed $\Delta = 1$, $T_i = 1/2$, $\varepsilon_1 = 1 / b_1$ and $\varepsilon_2 = 2 C / b_2$. Thus, by expanding the definitions of $b_1$ and $b_2$ as a function of $b$ and $C$, we can verify that \Cref{alg:SVT} is $(\varepsilon,0)$-DP with
\begin{align*}
\varepsilon &= \varepsilon_1 + \varepsilon_2 \\
&= \frac{1}{b_1} + \frac{2 C}{b_2} \\
&= \frac{1 + (2C)^{1/3}}{b} + \frac{(2C)^{2/3} (1 + 2C)^{1/3}}{b} \\
&= \frac{(1 + (2C)^{1/3}) (1 + (2C)^{2/3})}{b}.
\end{align*}
This concludes the proof.  \pagebreak

\section{Random Sampling Distributions} \label{sec:rsdists}
\Cref{tab:mnist-sampling-dist,tab:adult-sampling-dist} list the distributions for the hyperparameters used in the \MNIST{} and \Adult{} experiments respectively.

\begin{table}[h]
\centering
\adjustbox{max width=\columnwidth}{%
	\begin{tabular}{@{}lllll@{}}
		\toprule
		\textbf{Hyperparameter} & \textbf{Distribution} & \textbf{Parameters}       & \textbf{Int-Valued}       & \textbf{Accept Range} \\ \midrule
		Epochs         & Uniform            & $a = 1, b = 400$  & True   & $[1,400]$ \\
		Lot Size        & Normal 	& $\mu = 800, \sigma = 800$  & True   & $[16, 4000]$ \\
		Learning Rate         & Shifted Exp.            & $\lambda=10, \text{shift}=\num{1e-3}$  & False   & $[\num{1e-3}, \num{5e-1}]$ \\
		Noise Variance           & Shifted Exp.  &  $\lambda=\num{5e-1}, \text{shift}=\num{1e-1}$ & False & $[\num{1e-1}, 16]$ \\
		Clipping Norm           & Shifted Exp.  &  $\lambda=\num{5e-1}, \text{shift}=\num{1e-1}$ & False & $[\num{1e-1}, 12]$ \\ \bottomrule
	\end{tabular}}
	\caption{\MNIST{} random sampling distributions.}
	\label{tab:mnist-sampling-dist}
\end{table}

\begin{table}[h]
\centering
\adjustbox{max width=\columnwidth}{%
	\begin{tabular}{@{}lllll@{}}
		\toprule
		\textbf{Hyperparameter} & \textbf{Distribution} & \textbf{Parameters}       & \textbf{Int-Valued}       & \textbf{Accept Range} \\ \midrule
		Epochs         & Uniform            & $a = 1, b = 64$  & True   & $[1,64]$ \\
		Lot Size        & Normal 	& $\mu = 128, \sigma = 64$  & True   & $[8, 512]$ \\
		Learning Rate         & Shifted Exp.            & $\lambda=10, \text{shift}=\num{1e-3}$  & False   & $[\num{1e-3}, \num{1e-1}]$ \\
		Noise Variance          & Shifted Exp.  &  $\lambda=\num{1e-1}, \text{shift}=\num{1e-1}$ & False & $[\num{1e-1}, 16]$ \\
		Clipping Norm           & Shifted Exp.  &  $\lambda=\num{1e-1}, \text{shift}=\num{1e-1}$ & False & $[\num{1e-1}, 4]$ \\ \bottomrule
	\end{tabular}}
	\caption{\Adult{} random sampling distributions.}
	\label{tab:adult-sampling-dist}
\end{table}

\end{document}